%% file: MAIN.tex
\documentclass[letterpaper, 10 pt, conference]{class/ieeeconf}  

\IEEEoverridecommandlockouts                              
\usepackage{xcolor}
\usepackage{siunitx}  
\usepackage{graphicx}
 \graphicspath{./figures/robot_demonstration}

\usepackage{amsmath}
\usepackage{amssymb}
\usepackage{latexsym}
\usepackage{url}
\usepackage{cite}
\usepackage{relsize}
\usepackage{multirow}
\usepackage{afterpage}
\usepackage{ifthen}
\usepackage{graphicx}
\usepackage{algpseudocode,algorithm,algorithmicx}
\usepackage{subfigure}
\usepackage{flushend}
\usepackage{epstopdf}
\usepackage{stackengine}
\usepackage{textcomp} 

\usepackage{gensymb}
\usepackage{booktabs}
\usepackage{makecell}
\usepackage{hhline}
\usepackage{courier}
\usepackage{lipsum}
\usepackage{xspace}
\usepackage{xcolor} 
\usepackage{array} 

\usepackage{lpic}
            
\makeatletter
\let\NAT@parse\undefined
\makeatother
\usepackage[bookmarks=false, linkcolor=blue, urlcolor=blue, citecolor=blue]{hyperref} 
\hypersetup{
    colorlinks=true,
    linkcolor=red,
    filecolor=magenta,      
    urlcolor=blue,
    pdfstartview={FitH},
    citecolor =blue
    }

\newcommand{\second}[1]{\textit{#1}}
\usepackage{xspace}
\usepackage{import}
\usepackage{xcolor}
\usepackage{graphicx}
\usepackage[cal=cm,scr=euler]{mathalpha}
\usepackage{transparent}

 \graphicspath{{./figures/}}

\title{\LARGE \bf Lang2Lift: A Language-Guided Autonomous Forklift System for Outdoor Industrial Pallet Handling}

\author{Huy Hoang Nguyen$^{1}$, Johannes Huemer$^{1}$,  Markus Murschitz$^{1}$, Tobias Gl\"uck$^{1}$,\\ Minh Nhat Vu$^{2}$, Andreas Kugi$^{1,2}$ 
\thanks{$^1$ AIT Austrian Institute of Technology GmbH, Austria
}
\thanks{$^2$ Automation \& Control Institute, TU Wien, Austria 
}
}

\begin{document}
\input{0_localmacros}

\maketitle
\thispagestyle{empty}
\pagestyle{empty}

\begin{abstract}

Automating pallet handling in outdoor logistics and construction environments remains challenging due to unstructured scenes, variable pallet configurations, and changing environmental conditions. In this paper, we present \textit{Lang2Lift}, an end-to-end language-guided autonomous forklift system designed to support practical pallet pick-up operations in real-world outdoor settings. The system enables operators to specify target pallets using natural language instructions, allowing flexible selection among multiple pallets with different loads and spatial arrangements. Lang2Lift integrates foundation-model-based perception modules with motion planning and control in a closed-loop autonomy pipeline. Language-grounded visual perception is used to identify and segment target pallets, followed by 6D pose estimation and geometric refinement to generate manipulation-feasible insertion poses. The resulting pose estimates are directly coupled with the forklift’s planning and control modules to execute fully autonomous pallet pick-up maneuvers. We deploy and evaluate the proposed system on the ADAPT autonomous outdoor forklift platform across diverse real-world scenarios, including cluttered scenes, variable lighting, and different payload configurations. On a prompt-conditioned outdoor dataset (129 images, 387 prompt--image pairs), the proposed perception pipeline achieves consistent segmentation performance across challenging conditions, with mean IoU up to 0.59 and success rates exceeding 60\% at IoU$\ge$0.5 in the best configuration, while ablation results confirm the importance of mask refinement for precise pallet boundaries. Tolerance-based pose evaluation further indicates accuracy sufficient for successful fork insertion. Timing and failure analyses highlight key deployment trade-offs and practical limitations, providing insights into integrating language-guided perception within industrial automation systems. Video demonstrations are available at \href{https://eric-nguyen1402.github.io/lang2lift.github.io/}{\texttt{lang2lift.github.io}}.

\end{abstract}


\input{1_intro}

\input{2_rw} 
\input{3_method}

\input{4_exp}
\input{5_conclusions}

\bibliographystyle{class/IEEEtran}
\bibliography{class/IEEEabrv,class/reference}
   
\end{document}

%% file: 0_localmacros.tex

\newtheorem{problem}{Problem}
\newtheorem{lemma}{Lemma}
\newtheorem{theorem}[lemma]{Theorem}
\newtheorem{claim}{Claim}
\newtheorem{corollary}[lemma]{Corollary}
\newtheorem{definition}[lemma]{Definition}
\newtheorem{proposition}[lemma]{Proposition}
\newtheorem{remark}[lemma]{Remark}
\newenvironment{LabeledProof}[1]{\noindent{\it Proof of #1: }}{\qed}

\def\beq#1\eeq{\begin{equation}#1\end{equation}}
\def\bea#1\eea{\begin{align}#1\end{align}}
\def\beg#1\eeg{\begin{gather}#1\end{gather}}
\def\beqs#1\eeqs{\begin{equation*}#1\end{equation*}}
\def\beas#1\eeas{\begin{align*}#1\end{align*}}
\def\begs#1\eegs{\begin{gather*}#1\end{gather*}}

\newcommand{\poly}{\mathrm{poly}}
\newcommand{\eps}{\epsilon}
\newcommand{\e}{\epsilon}
\newcommand{\polylog}{\mathrm{polylog}}
\newcommand{\rob}[1]{\left( #1 \right)} 
\newcommand{\sqb}[1]{\left[ #1 \right]} 
\newcommand{\cub}[1]{\left\{ #1 \right\} } 
\newcommand{\rb}[1]{\left( #1 \right)} 
\newcommand{\abs}[1]{\left| #1 \right|} 
\newcommand{\zo}{\{0, 1\}}
\newcommand{\zonzo}{\zo^n \to \zo}
\newcommand{\zokzo}{\zo^k \to \zo}
\newcommand{\zot}{\{0,1,2\}}
\newcommand{\en}[1]{\marginpar{\textbf{#1}}}
\newcommand{\efn}[1]{\footnote{\textbf{#1}}}
\newcommand{\vecbm}[1]{\boldmath{#1}} 
\newcommand{\uvec}[1]{\hat{\vec{#1}}}
\newcommand{\thv}{\vecbm{\theta}}
\newcommand{\junk}[1]{}
\newcommand{\var}{\mathop{\mathrm{var}}}
\newcommand{\rank}{\mathop{\mathrm{rank}}}
\newcommand{\diag}{\mathop{\mathrm{diag}}}
\newcommand{\tr}{\mathop{\mathrm{tr}}}
\newcommand{\acos}{\mathop{\mathrm{acos}}}
\newcommand{\atantwo}{\mathop{\mathrm{atan2}}}
\newcommand{\SVD}{\mathop{\mathrm{SVD}}}
\newcommand{\quadf}{\mathop{\mathrm{q}}}
\newcommand{\linterp}{\mathop{\mathrm{l}}}
\newcommand{\sgn}{\mathop{\mathrm{sign}}}
\newcommand{\sym}{\mathop{\mathrm{sym}}}
\newcommand{\avg}{\mathop{\mathrm{avg}}}
\newcommand{\mean}{\mathop{\mathrm{mean}}}
\newcommand{\erf}{\mathop{\mathrm{erf}}}
\newcommand{\grad}{\nabla}
\newcommand{\R}{\mathbb{R}}
\newcommand{\defeq}{\triangleq}
\newcommand{\dims}[2]{[#1\!\times\!#2]}
\newcommand{\sdims}[2]{\mathsmaller{#1\!\times\!#2}}
\newcommand{\udims}[3]{#1}
\newcommand{\udimst}[4]{#1}
\newcommand{\com}[1]{\rhd\text{\emph{#1}}}
\newcommand{\ind}{\hspace{1em}}
\newcommand{\argmin}[1]{\underset{#1}{\operatorname{argmin}}}
\newcommand{\floor}[1]{\left\lfloor{#1}\right\rfloor}
\newcommand{\step}[1]{\vspace{0.5em}\noindent{#1}}
\newcommand{\quat}[1]{\ensuremath{\mathring{\mathbf{#1}}}}
\newcommand{\norm}[1]{\left\lVert#1\right\rVert}
\newcommand{\ignore}[1]{}
\newcommand{\specialcell}[2][c]{\begin{tabular}[#1]{@{}c@{}}#2\end{tabular}}
\newcommand*\Let[2]{\State #1 $\gets$ #2}
\newcommand{\algorithmicbreak}{\textbf{break}}
\newcommand{\Break}{\State \algorithmicbreak}
\newcommand{\ra}[1]{\renewcommand{\arraystretch}{#1}}

\renewcommand{\vec}[1]{\mathbf{#1}} 

\algdef{S}[FOR]{ForEach}[1]{\algorithmicforeach\ #1\ \algorithmicdo}
\algnewcommand\algorithmicforeach{\textbf{for each}}
\algrenewcommand\algorithmicrequire{\textbf{Require:}}
\algrenewcommand\algorithmicensure{\textbf{Ensure:}}
\algnewcommand\algorithmicinput{\textbf{Input:}}
\algnewcommand\INPUT{\item[\algorithmicinput]}
\algnewcommand\algorithmicoutput{\textbf{Output:}}
\algnewcommand\OUTPUT{\item[\algorithmicoutput]}

%% file: 1_intro.tex
\section{INTRODUCTION} \label{Sec:Intro}

Autonomous manipulation with forklifts in outdoor areas has attracted significant attention~\cite{iinuma2020pallet,syu2017computer,Teller2010avoice}, driven by skilled operator shortages and the critical need for increased efficiency in autonomous forklift operations~\cite{mohammadpour2024energy,BHAT2023anadvanced}. Current automated systems rely on rigid, preprogrammed characteristics, severely limiting adaptability when faced with new pallet types, unexpected orientations, or cluttered scenes. This rigidity forces sites to revert to manual operations or extensively reprogram their systems, resulting in increased costs, project delays, and safety risks.

Unlike controlled warehouse settings, outdoor operations demand systems that can interpret contextual requirements, such as "pick up the steel beam pallet near the crane" or "pick up the concrete block stack on the left", without pre-programmed knowledge of every pallet configuration. Thus, autonomous forklift operations in outdoor construction and logistics environments face a critical challenge: dynamically selecting and manipulating specific pallets from cluttered scenes containing various cargo types, orientations, and environmental conditions. This capability enables dynamic context-aware selection, providing the operational flexibility that human operators naturally possess in the pick-up operation. 

\begin{figure}
    \centering
    \includegraphics[scale=0.34]{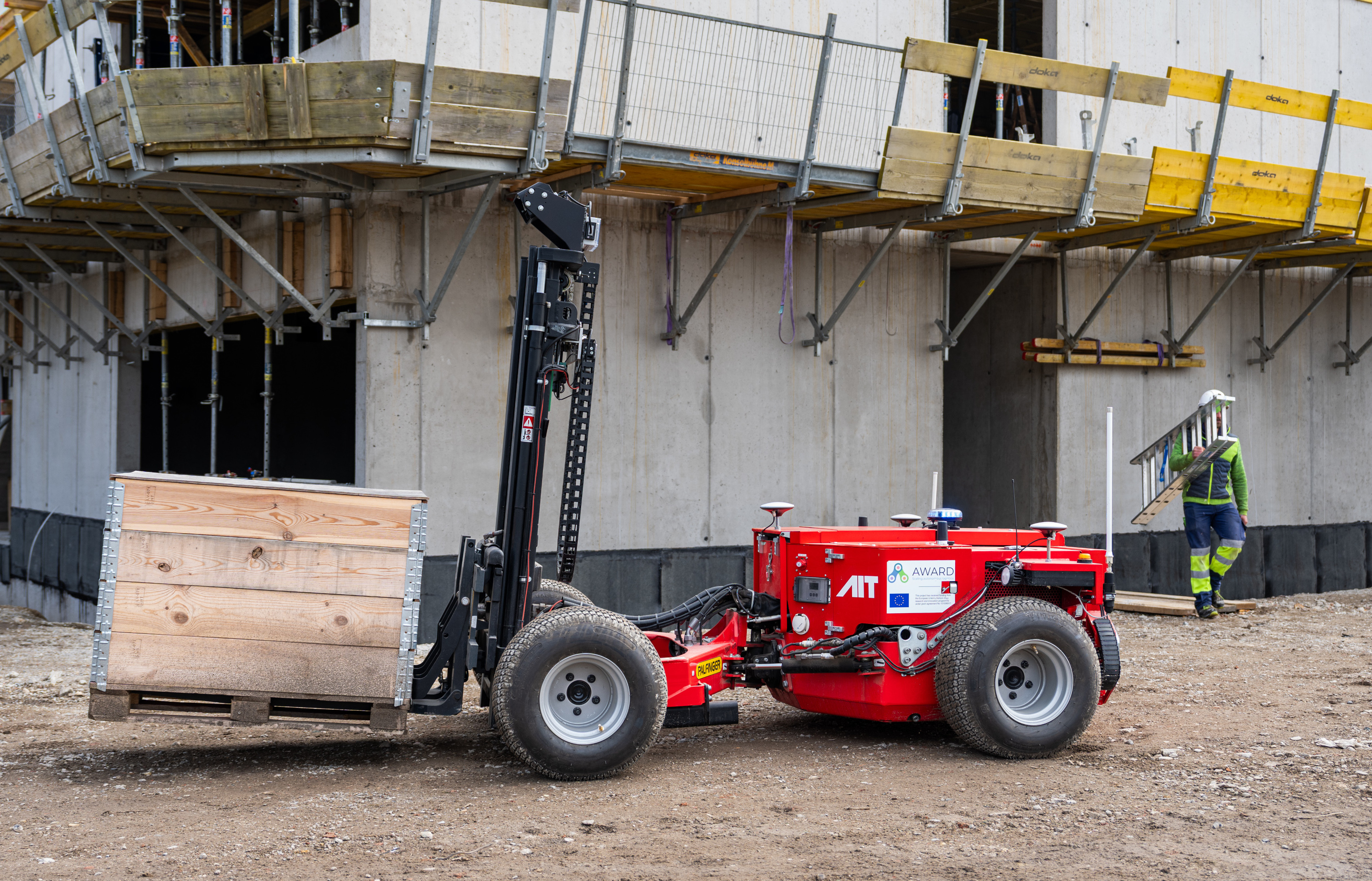}
    \vspace{0.2ex}
    \caption{The ADAPT autonomous outdoor forklift equipped with our Lang2Lift framework operating in outdoor conditions.}
    \label{fig:forklift}
\end{figure}

Although existing research has advanced pallet detection and localization~\cite{xiao2017pallet,tsiogas2021pallet, Rocha2023optimized, csaba2025pallet, beleznai2023automated}, these methods focus on detecting pallets rather than distinguishing between pallets carrying different types of loads. Thus, current approaches lack end-to-end solutions for language-guided pallet selection that can differentiate between various cargo configurations. While CNN-based methods~\cite{li2021pallet,mohamed2019detection,zaccaria2020comparison,caldana2024pallet} demonstrate reasonable performance on specific datasets for basic pallet detection, their performance often degrades under diverse outdoor conditions and cannot reliably classify the nature or presence of cargo loads. A remaining challenge is transitioning from rigid geometric pallet detection to flexible, context-aware systems capable of understanding and selecting pallets based on their load characteristics in response to natural language instructions.

Recent advances in foundation models (FM) have enabled new opportunities for integrating semantic perception into automation systems. These large-scale neural networks~\cite{xiao2023florence2,peng2024grounding} capture detailed semantic representations adaptable for downstream tasks~\cite{ravi2024sam2segmentimages}. Vision-language models (VLMs) present opportunities to create flexible robotic systems capable of outdoor operation. Natural language interfaces enable intuitive voice commands~\cite{park2024natural} that fundamentally bridge the human-robot communication gap by transforming rigid, preprogrammed systems into conversational interfaces. This transformation enables construction personnel to communicate with autonomous systems using familiar, natural language, rather than requiring technical programming expertise. It allows for real-time adaptability where human expertise can guide autonomous decisions through contextual commands. However, the integration of such models into industrial systems raises practical challenges related to latency, robustness, and deployment constraints.

This work focuses on system integration and real-world deployment rather than proposing new learning algorithms. Our work presents \textbf{Lang2Lift}, a framework integrating training-free foundation model modules for outdoor logistics perception. By combining natural language guidance with robust vision foundation models, \textbf{Lang2Lift} enables autonomous forklifts to handle real-world complexity while maintaining operational flexibility through intuitive human interaction. \textbf{Lang2Lift} is successfully deployed on the ADAPT autonomous outdoor forklift, Fig.~\ref{fig:forklift}, operating effectively in challenging field conditions. 

Our key contributions are summarized as follows:
\begin{itemize}
    \item We present an end-to-end language-guided autonomous forklift system that enables flexible pallet selection and pickup in outdoor industrial environments, and demonstrate its deployment on a full-scale autonomous forklift platform.

    \item We describe the practical integration of foundation-model-based perception with motion planning and control in a closed-loop autonomy pipeline, highlighting engineering considerations for real-world operation rather than algorithmic novelty.

    \item We provide a tolerance-driven quantitative evaluation that directly links perception and pose estimation accuracy to the feasibility of autonomous pallet manipulation.

    \item We analyze system timing, failure cases, and deployment limitations observed during real-world operation, offering insights into the challenges of applying language-guided perception in industrial automation settings.
\end{itemize}
    
    
    

%% file: 2_rw.tex
\section{Related Work} \label{Sec:rw}

Autonomous forklift operations have gained attention due to labor shortages and safety needs in material handling. Existing work focuses on specific aspects, such as pallet manipulation in outdoor settings~\cite{iinuma2020pallet, walter2010closedloop}. While \cite{walter2010closedloop} combines LiDAR, perception algorithms, and motion controllers for semi-structured environments, and \cite{iinuma2020pallet} demonstrates outdoor feasibility, these approaches rely on specific sensor configurations and controlled environments, limiting adaptability to diverse construction and logistics scenarios.

\subsection{Perception for Forklift Autonomy}

Perception capabilities fundamentally limit the effectiveness of autonomous systems in outdoor environments. Several researchers have investigated object segmentation and pose estimation using cameras or LiDAR~\cite{zhou2025pallet,shao2022novel,li2023systematic}. Current CNN-based methods~\cite{li2021pallet,mohamed2019detection,zaccaria2020comparison,caldana2024pallet} train models using user-defined datasets collected in real-world environments, usually achieving mIoU scores of 0.6-0.7 in controlled indoor settings. However, the lack of diverse and extensive training examples can lead to degraded generalization and reduced accuracy under varying outdoor conditions, with performance dropping to 0.4-0.5 mIoU~\cite{wuanewpallet} in challenging weather and lighting scenarios due to the limited diversity of the training data.
To address these limitations, language-driven approaches to object detection have emerged, allowing machines to understand and respond to complex, human-readable commands~\cite{yuan2024open,wu2024general,peng2024grounding}. Models like GroundingDINO~\cite{liu2023grounding} integrate vision and language by learning language-aware region embeddings, enabling open-vocabulary detection from textual input. However, these approaches have been validated primarily in controlled indoor environments with limited evaluation of their robustness to outdoor industrial conditions. Traditional perception methods often require large, labeled datasets and struggle with unstructured settings where variations in lighting, occlusion, and object clustering are common.

\subsection{Vision-Language Models in Robotics}

Foundation models like Florence-2~\cite{xiao2023florence2} offer unified capabilities for object detection, segmentation, and visual question answering through prompt engineering. Florence-2 demonstrates strong text-prompted object detection and visual grounding capabilities. Recent work employs SAM-2~\cite{ravi2024sam2segmentimages} to refine initial bounding boxes with fine-grained segmentation, achieving real-time performance under 200ms per frame.
Current applications face critical challenges: limited harsh outdoor evaluation, and a lack of integration with specialized pose estimation for precise manipulation. Most of the work focuses on controlled indoor conditions rather than challenging outdoor logistics conditions. Motivated by these challenges, our work explores the integration of vision-language models with a dedicated 6D pose estimation module within an autonomous forklift system.

\subsection{Object Pose Estimation}

Object pose estimation approaches can be categorized into model-based and model-free concepts. CAD model-based methods~\cite{labbe2022megapose,He2021ffb6d} achieve high accuracy, but require 3D models for each object, limiting scalability. Model-free methods~\cite{wen2023bundlesdf,he2022fs6dfewshot6dpose} offer flexibility but struggle with untextured objects and occlusions.
Foundation models represent a paradigm shift toward generalizable systems. FoundationPose~\cite{wen2024foundationpose} combines the strengths of model-based and model-free approaches for novel object pose estimation without fine-tuning, leveraging large-scale pre-training for improved generalization.
To the best of our knowledge, existing work has not reported integration of vision-language models with specialized pose estimation foundation models for real-time outdoor forklift operations, particularly in outdoor environments where natural language guidance is essential for task flexibility.

\begin{figure*}[!ht]
	\centering	
  \def\svgwidth{\textwidth}
   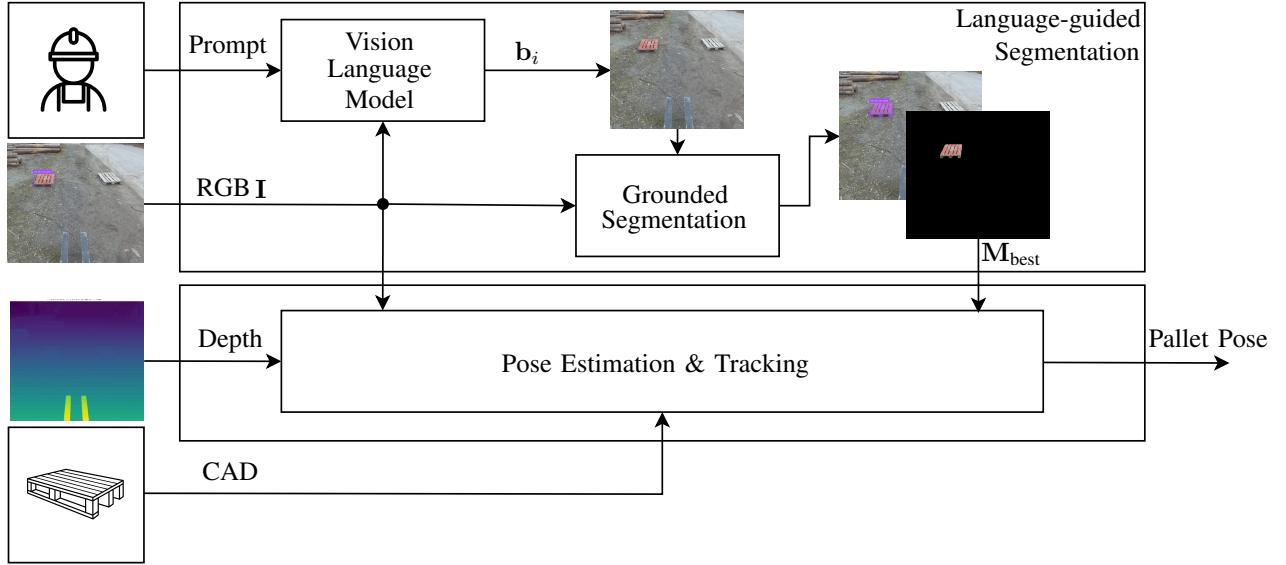
   \vspace{0.05ex}
	\caption{The \textbf{Lang2Lift} perception pipeline for automated pallet handling operations. The system processes natural language commands through Florence-2 for grounded object detection, applies SAM-2 for precise segmentation, and utilizes FoundationPose for 6D pose estimation with geometric refinement for optimal fork insertion positioning.}
	\label{fig:lang2lifts_flowchart}
\end{figure*}

%% file: figures/lang2lift_pipeline.pdf_tex
\begingroup%
  \makeatletter%
  \providecommand\color[2][]{%
    \errmessage{(Inkscape) Color is used for the text in Inkscape, but the package 'color.sty' is not loaded}%
    \renewcommand\color[2][]{}%
  }%
  \providecommand\transparent[1]{%
    \errmessage{(Inkscape) Transparency is used (non-zero) for the text in Inkscape, but the package 'transparent.sty' is not loaded}%
    \renewcommand\transparent[1]{}%
  }%
  \providecommand\rotatebox[2]{#2}%
  \newcommand*\fsize{\dimexpr\f@size pt\relax}%
  \newcommand*\lineheight[1]{\fontsize{\fsize}{#1\fsize}\selectfont}%
  \ifx\svgwidth\undefined%
    \setlength{\unitlength}{474.46028101bp}%
    \ifx\svgscale\undefined%
      \relax%
    \else%
      \setlength{\unitlength}{\unitlength * \real{\svgscale}}%
    \fi%
  \else%
    \setlength{\unitlength}{\svgwidth}%
  \fi%
  \global\let\svgwidth\undefined%
  \global\let\svgscale\undefined%
  \makeatother%
  \begin{picture}(1,0.27175997)%
    \lineheight{1}%
    \setlength\tabcolsep{0pt}%
    \put(0,0){\includegraphics[width=\unitlength,page=1]{lang2lift_pipeline.pdf}}%
    \put(0.14042105,0.23325064){\color[rgb]{0,0,0}\makebox(0,0)[t]{\lineheight{1.25}\smash{\begin{tabular}[t]{c}\scalebox{1}{Prompt}\end{tabular}}}}%
    \put(0.26146314,0.23921501){\color[rgb]{0,0,0}\makebox(0,0)[t]{\lineheight{1.25}\smash{\begin{tabular}[t]{c}\scalebox{1}{Vision-}\end{tabular}}}}%
    \put(0.26866242,0.21707358){\color[rgb]{0,0,0}\makebox(0,0)[t]{\lineheight{1.25}\smash{\begin{tabular}[t]{c}\scalebox{1}{Language}\end{tabular}}}}%
    \put(0.26075125,0.19172997){\color[rgb]{0,0,0}\makebox(0,0)[t]{\lineheight{1.25}\smash{\begin{tabular}[t]{c}\scalebox{1}{Model}\end{tabular}}}}%
    \put(0.50139839,0.14647623){\color[rgb]{0,0,0}\makebox(0,0)[t]{\lineheight{1.25}\smash{\begin{tabular}[t]{c}\scalebox{1}{Grounded}\end{tabular}}}}%
    \put(0.93643939,0.03571804){\color[rgb]{0,0,0}\makebox(0,0)[t]{\lineheight{1.25}\smash{\begin{tabular}[t]{c}\scalebox{1}{Pallet Pose}\end{tabular}}}}%
    \put(0.73191618,0.08935326){\color[rgb]{0,0,0}\makebox(0,0)[t]{\lineheight{1.25}\smash{\begin{tabular}[t]{c}\scalebox{1}{$\mathbf{M}_{\text{best}}$}\end{tabular}}}}%
    \put(0.72045592,0.24782538){\color[rgb]{0,0,0}\makebox(0,0)[t]{\lineheight{1.25}\smash{\begin{tabular}[t]{c}\scalebox{1}{Language-guided Segmentation}\end{tabular}}}}%
    \put(0.39735724,0.23467138){\color[rgb]{0,0,0}\makebox(0,0)[t]{\lineheight{1.25}\smash{\begin{tabular}[t]{c}\scalebox{1}{$\mathbf{b}_i$}\end{tabular}}}}%
    \put(0.50096152,0.11864013){\color[rgb]{0,0,0}\makebox(0,0)[t]{\lineheight{1.25}\smash{\begin{tabular}[t]{c}\scalebox{1}{Segmentation}\end{tabular}}}}%
    \put(0.50551654,0.03583326){\color[rgb]{0,0,0}\makebox(0,0)[t]{\lineheight{1.25}\smash{\begin{tabular}[t]{c}\scalebox{1}{Pose Estimation and Tracking}\end{tabular}}}}%
    \put(0.15969291,0.14578595){\color[rgb]{0,0,0}\makebox(0,0)[t]{\lineheight{1.25}\smash{\begin{tabular}[t]{c}\scalebox{1}{RGB I}\end{tabular}}}}%
    \put(0.14494321,0.04541601){\color[rgb]{0,0,0}\makebox(0,0)[t]{\lineheight{1.25}\smash{\begin{tabular}[t]{c}\scalebox{1}{Depth}\end{tabular}}}}%
    \put(0.84616944,0.21094465){\color[rgb]{0,0,0}\makebox(0,0)[t]{\lineheight{1.25}\smash{\begin{tabular}[t]{c}\scalebox{1}{CAD}\end{tabular}}}}%
    \put(0,0){\includegraphics[width=\unitlength,page=2]{lang2lift_pipeline.pdf}}%
  \end{picture}%
\endgroup%

%% file: 3_method.tex
\section{Lang2Lift Framework for Autonomous Forklifts}
\label{sec:methodology}

\subsection{\textbf{Lang2Lift} Perception Pipeline}

The perception pipeline, see Fig.~\ref{fig:lang2lifts_flowchart}, transforms natural language commands into actionable pose estimates through a three-stage process: language-driven object segmentation, pose estimation with geometric refinement, and temporal pose tracking. 

\subsubsection{Language-driven Object Segmentation}

The \textbf{Lang2Lift} perception pipeline begins with a language-driven object segmentation module that utilizes pretrained Vision FMs to recognize and localize pallets in outdoor construction and logistics environments without requiring task-specific training data.

\textbf{Natural Language Processing and Command Interpretation}: \textbf{Lang2Lift} processes natural language instructions through a lightweight semantic parsing module~\cite{walker2019neural} adapted for construction and logistics domains. Given a free-form command $C$, the parser extracts four key semantic components: \emph{object type} (e.g., “pallet”, “lumber pallet”), \emph{visual descriptors} (e.g., “red”, “with the concrete block on top”), \emph{spatial relationships} (e.g., “near the concrete mixer”, “behind the trailer”), and \emph{contextual references} (e.g., “left”, “closest to the crane”). Spatial relationships are mapped to a reference frame grounded in the forklift’s perception system using onboard sensor geometry.
The parsed command is encoded into a structured referring-expression prompt 
\begin{equation}
    T = \text{Format}(\text{type}, \text{descriptors}, \text{spatial}, \text{context}) \ ,
\end{equation}
which serves as direct input to the vision-language model. This approach enables operators to issue intuitive, high-level commands, such as:
\begin{itemize}
    \item \textit{"Pick up lumber pallet near the concrete mixer"}  
    \item \textit{"Pick up brick pallet stack behind the construction trailer"}
    \item \textit{"Pick up the pallet with the concrete block on top"}
\end{itemize}
By explicitly modeling task-specific language elements and mapping them to structured visual queries, our NLP module maintains real-time performance (average $<15$ ms per command) while preserving the descriptive richness needed for robust outdoor pallet detection.

\textbf{Vision Foundation Model-based Detection}: 
We employ Florence-2~\cite{xiao2023florence2}, a unified vision-language foundation model, for referring-expression-based object detection. Given an RGB image $\mathbf{I} \in \mathbb{R}^{H \times W \times 3}$ and a referring expression prompt $T$ derived from the natural language command, Florence-2 produces object detections
\begin{equation}
    \{\mathbf{b}_i, s_i\}_{i=1}^{N} = \text{Florence-2}(\mathbf{I}, T) \ ,
\end{equation}
where $\mathbf{b}^\top = [x_{\text{min}}, y_{\text{min}}, x_{\text{max}}, y_{\text{max}}]$ represents the bounding boxes, $s_i$ denotes confidence scores, and $N$ is the number of detected objects. 

\textbf{Fine-grained Segmentation with SAM-2}: To achieve the pixel-level accuracy required for precise pose estimation, we generate detailed segmentation masks using SAM-2~\cite{ravi2024sam2segmentimages}. For each detection bounding box $\mathbf{b}_i$ with confidence score $s_i > \theta_{\text{conf}}$, we apply SAM-2 to generate candidate segmentation masks

\begin{equation}
    \{\mathbf{M}_j, q_j, l_j\}_{j=1}^{K} = \text{SAM-2}(\mathbf{I}, \mathbf{b}_i) \ ,
\end{equation}
where $\mathbf{M}_j$ represents candidate masks, $q_j$ are quality scores, and $l_j$ are predicted logit scores. We select the mask with the highest quality score and convert it to a binary mask $\mathbf{M}_{\text{best}} \in \{0,1\}^{H \times W}$.

\subsubsection{Pose Processing Module}

The pose processing module estimates precise 6D object poses required for successful forklift manipulation, incorporating geometric constraints specific to pallet structures and outdoor operational requirements.

\textbf{Multi-modal Pose Estimation}: 
The system utilizes RGB-D data combined with segmentation masks to improve the robustness of pose estimation. Given the segmented image $\mathbf{I}$, corresponding depth information $\mathbf{D}$, and a pallet CAD model $\mathcal{M}$, we employ FoundationPose~\cite{wen2024foundationpose} to compute the initial 6D pose

\begin{equation}
    \mathbf{P}_{\text{init}} = [\mathbf{R} \mid \mathbf{t}] = \text{FoundationPose}(\mathbf{I}, \mathbf{D}, \mathbf{M}_{\text{best}}, \mathcal{M}) \ ,
\end{equation}
where $\mathbf{R} \in \text{SO}(3)$ represents the rotation matrix and $\mathbf{t} \in \mathbb{R}^3$ represents the translation vector in the camera coordinate frame. 

\begin{figure}[ht]
\vspace{0.05ex}
\centering
\def\svgwidth{1\columnwidth}
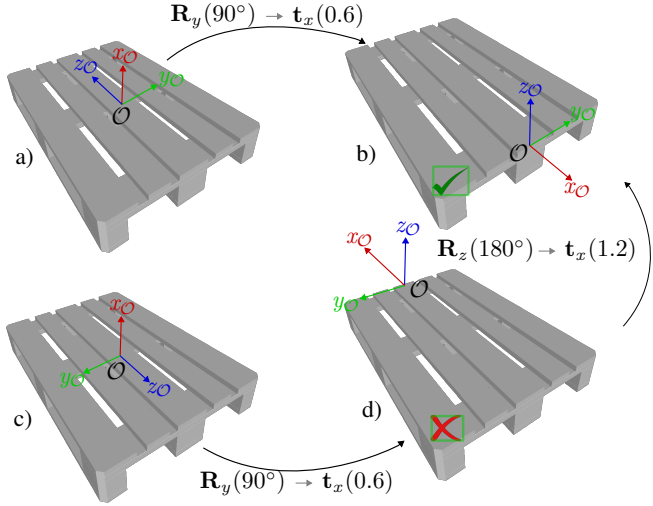
\vspace{0.05ex}
\caption{Pose transformation process showing: (a) initial pose detection with pallet symmetry creating two possible orientations, (b) target reference position for optimal fork insertion, and (c) alternative symmetric orientation requiring correction.}
\label{fig:pose_transform}
\end{figure}

\textbf{Geometric Refinement and Symmetry Handling}: 
The initial pose $\mathbf{P}_{\text{init}}$ is computed at the center coordinate of the pallet $\mathcal{O}$, following the FoundationPose convention. However, successful forklift operations require precise fork alignment with pallet pockets, which requires geometric transformation and resolution of symmetry. Due to pallet symmetry, FoundationPose yields two equally valid solutions corresponding to opposite insertion orientations (Fig.~\ref{fig:pose_transform}a and \ref{fig:pose_transform}c). To ensure consistent fork alignment, we implement a two-step geometric refinement process:

\begin{enumerate}
    \item \textbf{Orientation Disambiguation}: We determine the correct insertion orientation by evaluating the object's local $\mathit{x}$-axis direction relative to the camera position. The disambiguation criterion is
    \begin{equation}
        d = \mathbf{R}_x \cdot \mathbf{t} = R_{11} t_x + R_{12} t_y + R_{13} t_z \ ,
    \end{equation}
    where $\mathbf{R}_x^\top = [R_{11}, R_{12}, R_{13}]$ represents the first column of the rotation matrix $\mathbf{R}$.
    
    \item \textbf{Pose Transformation to Fork Reference Frame}: Based on the orientation check result, we apply the appropriate geometric transformation to position the reference frame at the optimal fork insertion point
    
    \begin{equation}
        \mathbf{P}_{\text{final}} = \begin{cases}
            \mathbf{T}_1 \cdot \mathbf{P}_{\text{init}} & \text{if } d > 0 \\
            \mathbf{T}_2 \cdot \mathbf{P}_{\text{init}} & \text{if } d \leq 0 \ ,
        \end{cases}
    \end{equation}
    where $\mathbf{T}_1$ (Fig.~\ref{fig:pose_transform}b) is a homogeneous transformation that represents a 90° clockwise rotation around the $\mathit{y}$-axis followed by a 0.6\,m translation along the $\mathit{x}$-axis, and $\mathbf{T}_2$ (Fig.~\ref{fig:pose_transform}d) includes an additional 180° rotation around the $\mathit{z}$-axis with a 1.2\,m $\mathit{x}$-axis translation.
\end{enumerate}

\subsubsection{Temporal Pose Tracking}
\label{pose_tracking}
To maintain robust pose estimates during dynamic forklift operations, \textbf{Lang2Lift} uses the temporal pose tracker from \cite{huemer2025adapt}. The system integrates vehicle odometry, GNSS measurements, and pallet detections from the perception pipeline into a unified probabilistic estimation framework operating at 25 Hz. The tracking architecture employs a factor graph-based approach that models forklift poses and pallet poses as variable nodes connected by measurement factors. Binary factors link pallet observations to the corresponding vehicle poses, whereas unary factors incorporate GNSS constraints and odometry measurements. 
Real-time optimization utilizes iSAM2~\cite{iSAM2} incremental smoothing, which maintains a dynamic Bayesian tree structure to efficiently update only affected graph portions when new measurements arrive. The system performs automatic marginalization of older poses beyond a 5-second window to maintain computational efficiency during extended operations.


\subsection{\textbf{Lang2Lift} Planning and Control Pipeline}

The planning and control pipeline transforms the perception system's pose estimates into safe and efficient forklift maneuvers through integrated motion planning and precise hydraulic control. This pipeline operates continuously to maintain real-time responsiveness while ensuring collision-free operation in dynamic outdoor environments.

\subsubsection{Motion Planning Architecture}

The \textbf{Lang2Lift} framework employs the hierarchical motion planning approach  from~\cite{huemer2025adapt} to deal with the unique challenges of articulated forklift vehicles in unstructured outdoor environments.

\textbf{Hybrid A* Path Planning}: 
Collision-free path planning utilizes an adapted Hybrid A* algorithm~\cite{dmitri2010pathplanning} specifically modified for articulated vehicle kinematics. The planner handles both forward and reverse maneuvers through Reeds-Shepp motion primitives, enabling efficient navigation in constrained spaces common to construction and logistics sites. The instantaneous centers of rotation are aligned using a car-like kinematic model that accounts for the articulated joint, enabling bi-directional maneuvers while maintaining stability constraints essential for safe forklift operation with loaded pallets.

\subsubsection{Vehicle Control and Fork Positioning}

\textbf{Path Following Control}:
The system implements a Lyapunov-based control law~\cite{bian2017kinematics} for robust path following that maintains stability under varying terrain conditions and load configurations. A multi-stage cascaded control architecture manages smooth transitions between forward and reverse movements by tracking a virtual reference vehicle and minimizing both lateral and heading errors.

\textbf{High-precision Fork Control}:
A dedicated fork control loop achieves centimeter-level positioning accuracy required for successful pallet engagement. The control system integrates real-time position feedback from fork-mounted sensors and vehicle odometry with precise PI(D) controllers for hydraulic valve spool adjustments, compensating for system nonlinearities and external disturbances.

\subsubsection{Task Planning and Execution}

A high-level task planner orchestrates the complete pallet manipulation sequence by coordinating perception updates, motion planning, and control execution. Additional implementation details for the autonomous navigation and pallet manipulation frameworks can be found in~\cite{huemer2025adapt}. The \textbf{Lang2Lift} framework integrates perception and planning via a ROS2-based architecture that manages data flow, timing, and fault recovery.







%% file: figures/pose_transfrom.pdf_tex
\begingroup%
  \makeatletter%
  \providecommand\color[2][]{%
    \errmessage{(Inkscape) Color is used for the text in Inkscape, but the package 'color.sty' is not loaded}%
    \renewcommand\color[2][]{}%
  }%
  \providecommand\transparent[1]{%
    \errmessage{(Inkscape) Transparency is used (non-zero) for the text in Inkscape, but the package 'transparent.sty' is not loaded}%
    \renewcommand\transparent[1]{}%
  }%
  \providecommand\rotatebox[2]{#2}%
  \newcommand*\fsize{\dimexpr\f@size pt\relax}%
  \newcommand*\lineheight[1]{\fontsize{\fsize}{#1\fsize}\selectfont}%
  \ifx\svgwidth\undefined%
    \setlength{\unitlength}{419.87336887bp}%
    \ifx\svgscale\undefined%
      \relax%
    \else%
      \setlength{\unitlength}{\unitlength * \real{\svgscale}}%
    \fi%
  \else%
    \setlength{\unitlength}{\svgwidth}%
  \fi%
  \global\let\svgwidth\undefined%
  \global\let\svgscale\undefined%
  \makeatother%
  \begin{picture}(1,0.765617)%
    \lineheight{1}%
    \setlength\tabcolsep{0pt}%
    \put(0,0){\includegraphics[width=\unitlength,page=1]{pose_transfrom.pdf}}%
    \put(0.1909455,0.68507492){\color[rgb]{0.78431373,0,0}\makebox(0,0)[t]{\lineheight{1.25}\smash{\begin{tabular}[t]{c}\scalebox{0.9}{$\mathit{x}_\mathcal{O}$}\end{tabular}}}}%
    \put(0.88703107,0.47851445){\color[rgb]{0.78431373,0,0}\makebox(0,0)[t]{\lineheight{1.25}\smash{\begin{tabular}[t]{c}\scalebox{0.9}{$\mathit{x}_\mathcal{O}$}\end{tabular}}}}%
    \put(0.26234269,0.6496976){\color[rgb]{0,0.79215686,0}\makebox(0,0)[t]{\lineheight{1.25}\smash{\begin{tabular}[t]{c}\scalebox{0.9}{$\mathit{y}_\mathcal{O}$}\end{tabular}}}}%
    \put(0.18489281,0.586024){\color[rgb]{0,0,0}\makebox(0,0)[t]{\lineheight{1.25}\smash{\begin{tabular}[t]{c}\scalebox{0.9}{$\mathcal{O}$}\end{tabular}}}}%
    \put(0.79625882,0.52782192){\color[rgb]{0,0,0}\makebox(0,0)[t]{\lineheight{1.25}\smash{\begin{tabular}[t]{c}\scalebox{0.9}{$\mathcal{O}$}\end{tabular}}}}%
    \put(0.89560384,0.59681163){\color[rgb]{0,0.79215686,0}\makebox(0,0)[t]{\lineheight{1.25}\smash{\begin{tabular}[t]{c}\scalebox{0.9}{$\mathit{y}_\mathcal{O}$}\end{tabular}}}}%
    \put(0.13036077,0.66933398){\color[rgb]{0,0,0.8627451}\makebox(0,0)[t]{\lineheight{1.25}\smash{\begin{tabular}[t]{c}\scalebox{0.9}{$\mathit{z}_\mathcal{O}$}\end{tabular}}}}%
    \put(0.8151872,0.63325559){\color[rgb]{0,0,0.8627451}\makebox(0,0)[t]{\lineheight{1.25}\smash{\begin{tabular}[t]{c}\scalebox{0.9}{$\mathit{z}_\mathcal{O}$}\end{tabular}}}}%
    \put(0,0){\includegraphics[width=\unitlength,page=2]{pose_transfrom.pdf}}%
    \put(0.45360362,0.06914713){\color[rgb]{0,0,0}\makebox(0,0)[t]{\lineheight{1.25}\smash{\begin{tabular}[t]{c}\scalebox{0.9}{$\mathbf{T}_1$}\end{tabular}}}}%
    \put(0,0){\includegraphics[width=\unitlength,page=3]{pose_transfrom.pdf}}%
    \put(0.18704237,0.29840231){\color[rgb]{0.78431373,0,0}\makebox(0,0)[t]{\lineheight{1.25}\smash{\begin{tabular}[t]{c}\scalebox{0.9}{$\mathit{x}_\mathcal{O}$}\end{tabular}}}}%
    \put(0.55161525,0.40280894){\color[rgb]{0.78431373,0,0}\makebox(0,0)[t]{\lineheight{1.25}\smash{\begin{tabular}[t]{c}\scalebox{0.9}{$\mathit{x}_\mathcal{O}$}\end{tabular}}}}%
    \put(0.10967009,0.18678798){\color[rgb]{0,0.79215686,0}\makebox(0,0)[t]{\lineheight{1.25}\smash{\begin{tabular}[t]{c}\scalebox{0.9}{$\mathit{y}_\mathcal{O}$}\end{tabular}}}}%
    \put(0.17680567,0.19226091){\color[rgb]{0,0,0}\makebox(0,0)[t]{\lineheight{1.25}\smash{\begin{tabular}[t]{c}\scalebox{0.9}{$\mathcal{O}$}\end{tabular}}}}%
    \put(0.64342777,0.31952856){\color[rgb]{0,0,0}\makebox(0,0)[t]{\lineheight{1.25}\smash{\begin{tabular}[t]{c}\scalebox{0.9}{$\mathcal{O}$}\end{tabular}}}}%
    \put(0.53077132,0.30313833){\color[rgb]{0,0.79215686,0}\makebox(0,0)[t]{\lineheight{1.25}\smash{\begin{tabular}[t]{c}\scalebox{0.9}{$\mathit{y}_\mathcal{O}$}\end{tabular}}}}%
    \put(0.24318162,0.17109622){\color[rgb]{0,0,0.8627451}\makebox(0,0)[t]{\lineheight{1.25}\smash{\begin{tabular}[t]{c}\scalebox{0.9}{$\mathit{z}_\mathcal{O}$}\end{tabular}}}}%
    \put(0.62661796,0.41939503){\color[rgb]{0,0,0.8627451}\makebox(0,0)[t]{\lineheight{1.25}\smash{\begin{tabular}[t]{c}\scalebox{0.9}{$\mathit{z}_\mathcal{O}$}\end{tabular}}}}%
    \put(0,0){\includegraphics[width=\unitlength,page=4]{pose_transfrom.pdf}}%
    \put(0.41572407,0.69612701){\color[rgb]{0,0,0}\makebox(0,0)[t]{\lineheight{1.25}\smash{\begin{tabular}[t]{c}\scalebox{0.9}{$\mathbf{T}_1$}\end{tabular}}}}%
    \put(0.94434799,0.37676158){\color[rgb]{0,0,0}\makebox(0,0)[t]{\lineheight{1.25}\smash{\begin{tabular}[t]{c}\scalebox{0.9}{$\mathbf{T}_2$}\end{tabular}}}}%
    \put(0.03597507,0.52351643){\color[rgb]{0,0,0}\makebox(0,0)[t]{\lineheight{1.25}\smash{\begin{tabular}[t]{c}\scalebox{0.9}{a)}\end{tabular}}}}%
    \put(0.03378367,0.11965016){\color[rgb]{0,0,0}\makebox(0,0)[t]{\lineheight{1.25}\smash{\begin{tabular}[t]{c}\scalebox{0.9}{c)}\end{tabular}}}}%
    \put(0.56231826,0.52846981){\color[rgb]{0,0,0}\makebox(0,0)[t]{\lineheight{1.25}\smash{\begin{tabular}[t]{c}\scalebox{0.9}{b)}\end{tabular}}}}%
    \put(0.57079268,0.13337222){\color[rgb]{0,0,0}\makebox(0,0)[t]{\lineheight{1.25}\smash{\begin{tabular}[t]{c}\scalebox{0.9}{d)}\end{tabular}}}}%
  \end{picture}%
\endgroup%

%% file: 4_exp.tex
\section{Experimental Results}
\label{sec:experimental_results}

\begin{figure}[!ht]
	\centering	
    \def\svgwidth{1\columnwidth}
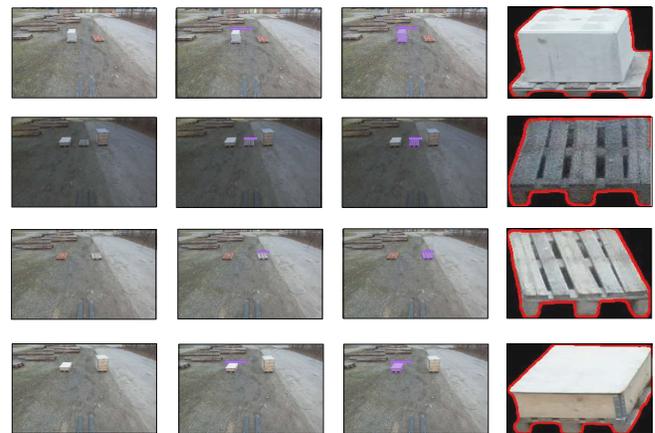
\vspace{0.05ex}
	\caption{Representative examples of successful pallet segmentation across diverse scenarios with corresponding natural language instructions, demonstrating robust performance under varying lighting conditions, load configurations, and spatial arrangements.}
	\label{fig:segmentation_examples}
\end{figure}

We conducted experiments in an outdoor laboratory for large-scale utility machinery testing using a truck-mounted, remote-controlled forklift platform and an autonomy stack~\cite{huemer2025adapt}. For this study, perception uses a ZED~2i RGB-D stereo camera, while an onboard LiDAR supports mapping and provides pose ground truth via manual point-cloud annotation after extrinsic calibration. Perception runs on a research-prototype workstation (Intel i9-14900K, 32\,GB RAM, RTX~4090 24\,GB), reflecting feasibility evaluation rather than embedded deployment. To validate robustness across scenarios, we test diverse pallet configurations (empty pallets, concrete blocks, wooden boxes; Fig.~\ref{fig:segmentation_examples}) and evaluate a dataset of 129 images with 387 human-generated prompts collected under sunny, snowy, low-light, and occlusion conditions. Images are from our test site and public non-AI internet sources; ground-truth masks are manually annotated in Roboflow, and pose ground truth is obtained from the calibrated LiDAR point clouds. Dataset details are available at \href{https://eric-nguyen1402.github.io/lang2lift.github.io/}{\texttt{lang2lift.github.io}}.

\subsection{Prompt-Conditioned Segmentation Performance Analysis}

\begin{table}[t]
    \centering
    \caption{Prompt-conditioned pallet segmentation on our outdoor forklift test set. For each prompt--image pair, the system outputs a single top-1 mask. We report mean IoU (mIoU) with 95\% confidence interval (CI), and success rate (SR) at IoU thresholds 0.5 and 0.75. \textbf{Best overall} values are in bold.}
    \label{tab:pallet-detection}
    \vspace{0.3em}
    \renewcommand{\arraystretch}{1.08}
    \resizebox{\columnwidth}{!}{%
    \begin{tabular}{>{\raggedright\arraybackslash}p{3.4cm} l r c c c}
        \toprule
        \textbf{Method} & \textbf{Condition} & \textbf{n} & \textbf{mIoU} $\uparrow$ & \textbf{SR@0.5} $\uparrow$ & \textbf{SR@0.75} $\uparrow$ \\
        \midrule
        \multirow{5}{=}{Lang2Lift (Florence-2 phrase grounding + SAM-2)} 
            & Sunny      & 126 & 0.398$\pm$0.076 & 37.30 & 35.71 \\
            & Snowy      & 123 & 0.458$\pm$0.074 & 47.97 & 38.21 \\
            & Low-light  & 36  & 0.805$\pm$0.107 & 83.33 & 83.33 \\
            & Occlusion  & 102 & 0.332$\pm$0.083 & 33.33 & 32.35 \\
            & Overall    & 387 & 0.437$\pm$0.044 & 43.93 & 40.05 \\
        \midrule
        \multirow{5}{=}{Lang2Lift (Florence-2 open-vocab + SAM-2) [language ablation]} 
            & Sunny      & 126 & 0.586$\pm$0.073 & 59.52 & 52.38 \\
            & Snowy      & 123 & 0.663$\pm$0.061 & 68.29 & 58.54 \\
            & Low-light  & 36  & 0.624$\pm$0.115 & 66.67 & 50.00 \\
            & Occlusion  & 102 & 0.481$\pm$0.085 & 50.00 & 47.06 \\
            & Overall    & 387 & \textbf{0.587$\pm$0.040} & \textbf{60.47} & \textbf{52.71} \\
        \midrule
        \multirow{5}{=}{GroundingDINO + SAM-2 [open-vocab baseline]} 
            & Sunny      & 126 & 0.504$\pm$0.073 & 47.62 & 45.24 \\
            & Snowy      & 123 & 0.620$\pm$0.068 & 63.41 & 60.98 \\
            & Low-light  & 36  & 0.800$\pm$0.090 & 91.67 & 75.00 \\
            & Occlusion  & 102 & 0.360$\pm$0.079 & 32.35 & 32.35 \\
            & Overall    & 387 & \second{0.531$\pm$0.041} & \second{52.71} & \second{49.61} \\
        \midrule
        \multirow{5}{=}{Florence-2 box-mask (no SAM-2) [segmentation ablation]} 
            & Sunny      & 126 & 0.341$\pm$0.047 & 30.95 & 7.14 \\
            & Snowy      & 123 & 0.438$\pm$0.040 & 53.66 & 2.44 \\
            & Low-light  & 36  & 0.425$\pm$0.082 & 41.67 & 16.67 \\
            & Occlusion  & 102 & 0.291$\pm$0.058 & 26.47 & 14.71 \\
            & Overall    & 387 & 0.367$\pm$0.027 & 37.98 & 8.53 \\
        \bottomrule
    \end{tabular}%
    }
    \vspace{0.2em}
    \footnotesize{CI is computed over prompt--image pairs using a normal approximation. SR values are in \% (higher is better).}
\end{table}

\begin{figure}[ht]
\centering
\def\svgwidth{1\columnwidth}
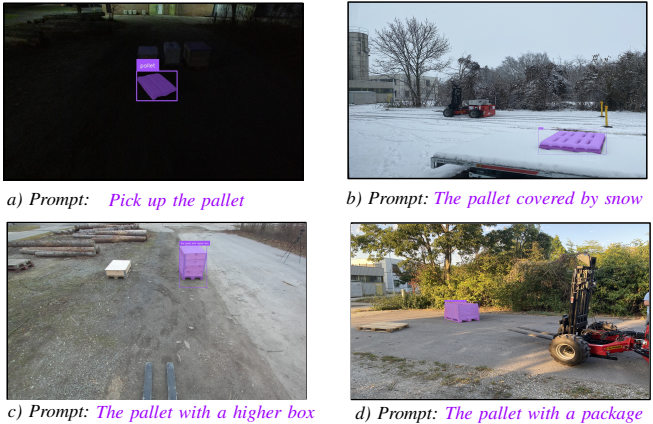
\vspace{0.02ex}
\caption{The visualization of vision-language object segmentation within different conditions.}
\label{fig:visual_detection}
\end{figure}

Table~\ref{tab:pallet-detection} reports prompt-conditioned pallet segmentation on our outdoor forklift test set (129 images, 387 prompt--image pairs; Fig.~\ref{fig:visual_detection}). 
Unlike conventional instance segmentation benchmarks, our evaluation is \emph{query-based}: each test sample is a (prompt, image) pair. For each pair, the system outputs a single top-1 pallet mask. We compute intersection-over-union (IoU) between the predicted mask and ground-truth pallet masks in the image, using the best-matching ground-truth instance for that query. We report mean IoU (mIoU) with a 95\% confidence interval (CI), and success rate (SR) at IoU thresholds 0.5 and 0.75 (the fraction of prompt--image pairs achieving IoU $\ge$ 0.5 and IoU $\ge$ 0.75, respectively).

\textbf{Comparison to open-vocabulary baseline.} We compare against a training-free open-vocabulary baseline using GroundingDINO for box grounding followed by SAM-2 for mask refinement. Overall, GroundingDINO+SAM-2 achieves mIoU 0.531 and SR@0.75 49.61\%, while our Florence-2 open-vocabulary variant (``pallet'' only + SAM-2) achieves higher overall accuracy (mIoU 0.587 and SR@0.75 52.71\%). Both approaches degrade under occlusion, which remains the most challenging regime (mIoU 0.360 for GroundingDINO+SAM-2; 0.481 for Florence-2 ``pallet'' only), highlighting the importance of robust downstream control policies and clearance margins in cluttered yards.

\textbf{Ablation on language prompting.} The open-vocabulary ``pallet'' query outperforms phrase grounding overall (mIoU 0.587 vs. 0.437). This suggests that on our current test distribution, where pallets are often visually salient, a fixed class query is frequently sufficient, while detailed natural-language referring expressions can introduce ambiguity (e.g., competing nearby objects or underspecified attributes). Notably, in low-light conditions, Lang2Lift achieves strong performance (mIoU 0.805, SR@0.75 83.33\%), indicating that descriptive prompts can be beneficial when photometric cues are weak, and context words help disambiguate the target.

\textbf{Ablation on mask refinement.} Removing SAM-2 and filling Florence-2 boxes as rectangular masks substantially reduces strict-overlap success (overall SR@0.75 drops to 8.53\%), confirming that SAM-2 provides essential boundary refinement for accurate pallet geometry, which is important for stable 6D pose estimation and safe fork insertion.

\textbf{Architectural choice: generalist vs. specialist grounding.} While Florence-2 and GroundingDINO provide comparable training-free performance when paired with SAM-2, we select Florence-2 as the core perception model because it is a \emph{generalist} foundation model that supports a broader set of vision tasks within a unified interface (e.g., open-vocabulary detection and phrase grounding within the same model family). This design choice increases pipeline flexibility for language-guided autonomy: the same perception backend can be reused for different operator intents (e.g., grounding, verification queries, or extended scene understanding) without introducing separate task-specific models. In practice, this reduces engineering overhead when expanding the system to additional language-guided manipulation objectives beyond pallet pickup.

\subsection{6D Pose Estimation Accuracy}

The objective of this evaluation is not to benchmark perception models against state-of-the-art methods, but to assess whether the resulting pose estimates meet the operational tolerances required for reliable autonomous forklift manipulation. Therefore, we refer to the results reported in the FoundationPose~\cite{wen2024foundationpose}, which establishes its state-of-the-art performance and highlights its unique capability as a unified framework for both model-based and model-free 6D pose estimation. On the YCB-Video benchmark~\cite{xiang2017posecnn} using established ADD and ADD-S (ADD-Symmetric) metrics for the 6D pose accuracy evaluation, the results presented in \cite{wen2024foundationpose} demonstrate state-of-the-art performance with our zero-shot foundation model approach, achieving an ADD of 0.91 and an ADD-S of 0.97. This represents a significant improvement over the previous best results reported by \cite{vu2024occlusion}, which achieved an ADD of 0.86 and an ADD-S of 0.92 using a specialized RGB-D pipeline with a deep learning–based method specifically tailored for pallet pose estimation.

\begin{figure}[ht]
    \centering
    \def\svgwidth{1\columnwidth}
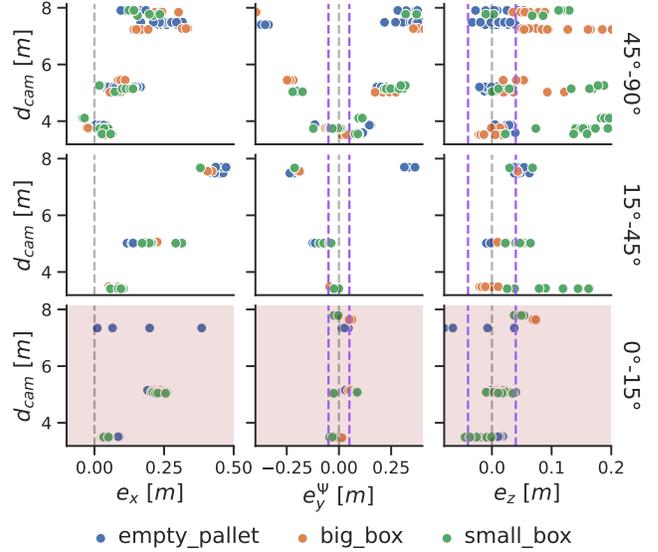
    \vspace{0.02ex}
    \caption{Quantitative analysis of pose estimation accuracy for pallet detection across varying operational parameters. The figure shows pallet pose estimation errors as a function of distance to the camera (\(d_{\text{cam}}\)) for various load types. Rows represent different orientations around the pallet’s $z$-axis. Dashed lines indicate tolerance boundaries for successful fork insertion, with the highlighted $0^\circ-15^\circ$ range being critical for insertion.}
    \label{fig:pallet_accuracy}
\end{figure}

\subsection{Tolerance Requirements for Successful Manipulation}

Autonomous pallet manipulation requires precise adherence to kinematic and geometric constraints of forklift parameters and pallet dimensions. Critical tolerance thresholds ensure reliable fork insertion: lateral accuracy within $\pm 0.05$~meters and vertical clearance of $\pm 0.04$~meters. Roll and pitch deviations are filtered to maintain ground-parallel assumptions.

\subsubsection{Methodology and Error Analysis Framework}

Tolerance specifications are derived from pallet geometry standards and forklift dimensional constraints, enabling quantitative assessment of pose estimation accuracy. Ground truth data was acquired through manual annotation of point cloud data from an extrinsically calibrated LiDAR sensor. The evaluation encompasses 284 detection instances in multiple load configurations, distances, and orientations (Fig.~\ref{fig:pallet_accuracy}). Each 6D pose estimate undergoes a coordinate transformation into a reference frame centered on the manual annotation (Fig.~\ref{fig:pose_transform}b), yielding error vector $\mathbf{e}^\top = [e_x, e_y, e_z, e_\phi, e_\theta, e_\psi]$. Operational tolerance limits are $y_{\text{tol}} = \pm 0.05$~meters and $z_{\text{tol}} = \pm 0.04$~meters in the pallet coordinate system, with $x$-direction constrained by load center-of-mass considerations.

\subsubsection{Angular Error Compensation and Filtering}

The evaluation methodology incorporates error compensation for angular deviations that affect the practical geometry of the insertion. Errors in the lateral direction ($e_y$) experience amplification due to orientation deviations around the $z$-axis of the pallet, which effectively modify the apparent opening dimension of the pallet by a factor of $\cos(e_\psi)$. This geometric relationship is integrated into the lateral error calculation as
$
e_y^{\psi} = e_y/\cos(e_\psi).
$
This formulation provides a more accurate representation of operational impact by accounting for the geometric coupling between angular and translational positioning errors.
Detections exhibiting unsuitable roll and pitch angles undergo systematic filtering, as forklift mechanical constraints prevent successful engagement of nonhorizontal pallets. Consequently, both rotational dimensions are excluded from the primary evaluation metrics while remaining available for system health monitoring and diagnostic purposes.

\subsubsection{Performance Characterization and Distance Dependencies}

The evaluation results demonstrate clear performance trends with pose accuracy degrading as sensor distance increases, particularly for longitudinal errors ($e_x$) due to quadratic depth error scaling in stereo vision. The error distribution shifts between the longitudinal components ($e_x$) and lateral ($e_y$) components with $z$-axis rotation, while load-dependent effects increase the vertical estimation errors ($e_z$) in extended ranges due to confusion of the pallet-cargo boundary. Single detection attempts at angles exceeding $15^\circ$ demonstrate insufficient reliability, although dynamic improvement during approach maneuvers consistently achieves operational tolerances for successful manipulation.

\subsection{Timing Analysis}


\begin{table}[ht]
\centering
\caption{Detailed Timing Analysis for Complete Autonomous Pallet Handling Maneuvers.}
\label{tab:average_timings}
\vspace{0.5em}
\begin{tabular}{|l|c|c|}
\hline
\textbf{Pipeline Component} & \textbf{Avg. Time (s)} & \textbf{Freq. (Hz)} \\ \hline
VLM Detection (Florence-2) & 0.14 & 7.1 \\ \hline
Grounded Segmentation (SAM-2) & 0.04 & 25.0 \\ \hline
Pose Estimation \& Adjustment & 0.83 & 1.2 \\ \hline
Pose Tracking  & 0.04& 25.0 \\ \hline
\textbf{Total Perception Pipeline Cycle} & \textbf{1.05} & \textbf{0.95} \\ \hline
Motion Planning & 0.40 & 2.5 \\ \hline
\textbf{Total Planning Pipeline Cycle} & \textbf{1.45} & \textbf{0.69} \\ \hline
\end{tabular}
\end{table}

The timing results in Table~\ref{tab:average_timings} can be interpreted in the context of the physical pallet pick-up sequence, from initial search to fork insertion and lift. The perception pipeline completes a full language-to-pose cycle in approximately 1.05\,s, while the combined perception and planning loop takes 1.45\,s on average. For outdoor forklift operation at low approach speeds, this rate is sufficient to maintain safe and accurate control, provided that high-rate submodules keep the scene state fresh between full cycles.

VLM detection runs at 0.14\,s, enabling rapid filtering and localization of pallets from natural-language commands-critical for target identification in cluttered scenes. This is followed by SAM-2 at 0.04\,s (25 Hz), refining region-of-interest masks to provide the 6D pose solver with accurate boundaries under varying lighting and clutter. The dominant latency arises from pose estimation and geometric adjustment at 0.83\,s (1.2 Hz), which, while sufficient during approach, limits how frequently precise insertion poses can be re-verified. High-rate pose tracking (25 Hz) mitigates this by maintaining smooth and accurate target estimates between slower 6D updates, ensuring fork placement within $\pm 0.05$ meters lateral and $\pm 0.04$ meters vertical tolerances during final alignment. Motion planning adds 0.40\,s (2.5 Hz) and, though currently sequential, could overlap with the latter part of pose estimation once an intermediate stable pose is available, reducing cycle time by 0.25–0.35\,s. Reducing pose estimation latency and enabling perception–planning overlap would tighten the perception-to-actuation loop, allowing more frequent pose re-verification and improving robustness in dynamic, unstructured environments. The reported timing results correspond to a research prototype configuration and are intended to evaluate system integration and operational feasibility rather than deployment on embedded hardware. Future work will investigate model compression, asynchronous perception–planning pipelines, and execution on edge-grade platforms, guided by the safety-critical nature and low-speed operation of industrial forklifts.

\subsection{Failure Case Analysis and System Limitations}

\begin{figure}[ht]
\centering
\includegraphics[width=1\columnwidth]{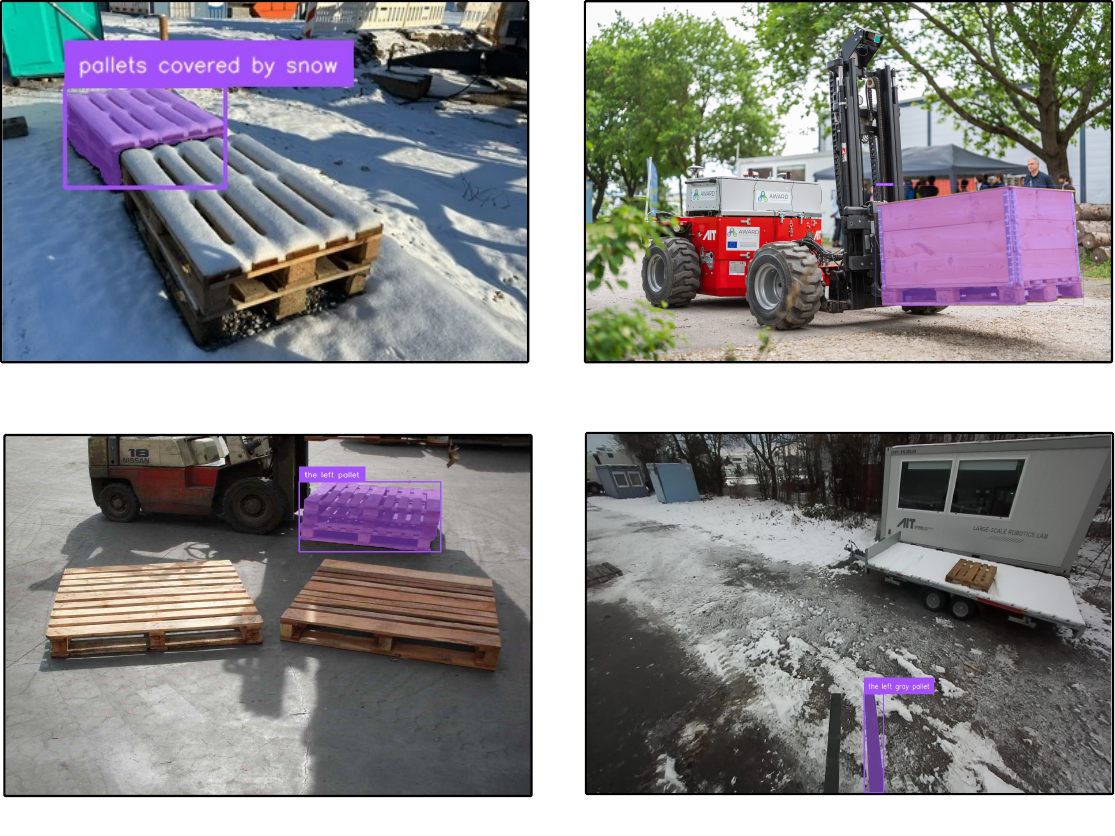}
\caption{Representative failure cases illustrating current system limitations in linguistic processing, complete occlusion scenarios, image quality sensitivity, and complex multi-object scene handling.}
\label{fig:failure_cases}
\end{figure}

Despite strong overall performance, we identified several failure modes that inform future development priorities. Fig.~\ref{fig:failure_cases} illustrates representative failure cases encountered during testing in various scenarios.
Linguistic processing challenges include sensitivity to grammatical variations, particularly in the distinction between singular and plural forms, which can lead to inconsistent object selection. Complex spatial relationships can result in ambiguous references, while insufficient contextual prompts reduce detection confidence. Visual processing limitations emerge primarily in complete occlusion scenarios, low image quality, and complex multi-object scenes, where dense clusters can lead to segmentation boundary errors.
These failure modes indicate important deployment considerations. Clear natural language command protocols should be established during operator training.

%% file: figures/examples_seg.pdf_tex
\begingroup%
  \makeatletter%
  \providecommand\color[2][]{%
    \errmessage{(Inkscape) Color is used for the text in Inkscape, but the package 'color.sty' is not loaded}%
    \renewcommand\color[2][]{}%
  }%
  \providecommand\transparent[1]{%
    \errmessage{(Inkscape) Transparency is used (non-zero) for the text in Inkscape, but the package 'transparent.sty' is not loaded}%
    \renewcommand\transparent[1]{}%
  }%
  \providecommand\rotatebox[2]{#2}%
  \newcommand*\fsize{\dimexpr\f@size pt\relax}%
  \newcommand*\lineheight[1]{\fontsize{\fsize}{#1\fsize}\selectfont}%
  \ifx\svgwidth\undefined%
    \setlength{\unitlength}{697.75657654bp}%
    \ifx\svgscale\undefined%
      \relax%
    \else%
      \setlength{\unitlength}{\unitlength * \real{\svgscale}}%
    \fi%
  \else%
    \setlength{\unitlength}{\svgwidth}%
  \fi%
  \global\let\svgwidth\undefined%
  \global\let\svgscale\undefined%
  \makeatother%
  \begin{picture}(1,0.52881844)%
    \lineheight{1}%
    \setlength\tabcolsep{0pt}%
    \put(0,0){\includegraphics[width=\unitlength,page=1]{examples_seg.pdf}}%
    \put(0.2448798,0.51706621){\color[rgb]{0,0,0}\makebox(0,0)[t]{\lineheight{1.25}\smash{\begin{tabular}[t]{c}\scalebox{0.6}{\textit{Prompt:}}\end{tabular}}}}%
    \put(0.3201439,0.3509113){\color[rgb]{0,0,0}\transparent{0.99644101}\makebox(0,0)[t]{\lineheight{1.25}\smash{\begin{tabular}[t]{c}\scalebox{0.6}{\textit{Prompt:}}\end{tabular}}}}%
    \put(0.53966882,0.34921853){\color[rgb]{0.61568627,0,1}\transparent{0.99644101}\makebox(0,0)[t]{\lineheight{1.25}\smash{\begin{tabular}[t]{c}\scalebox{0.6}{\textit{Pick up the pallet in the middle}}\end{tabular}}}}%
    \put(0.246028,0.17840035){\color[rgb]{0,0,0}\makebox(0,0)[t]{\lineheight{1.25}\smash{\begin{tabular}[t]{c}\scalebox{0.6}{\textit{Prompt:}}\end{tabular}}}}%
    \put(0.53266112,0.17737232){\color[rgb]{0.61568627,0,1}\makebox(0,0)[t]{\lineheight{1.25}\smash{\begin{tabular}[t]{c}\scalebox{0.6}{\textit{Pick up the left pallet with shorter box on top}}\end{tabular}}}}%
    \put(0.10671652,0.00286534){\color[rgb]{0,0,0}\makebox(0,0)[t]{\lineheight{1.25}\smash{\begin{tabular}[t]{c}\scalebox{0.6}{RGB Image}\end{tabular}}}}%
    \put(0.3678463,0.0036286){\color[rgb]{0,0,0}\makebox(0,0)[t]{\lineheight{1.25}\smash{\begin{tabular}[t]{c}\scalebox{0.6}{VLMs detection}\end{tabular}}}}%
    \put(0.62769783,0.00369116){\color[rgb]{0,0,0}\makebox(0,0)[t]{\lineheight{1.25}\smash{\begin{tabular}[t]{c}\scalebox{0.6}{Grounded Segmentation}\end{tabular}}}}%
    \put(0.88086291,0.00427641){\color[rgb]{0,0,0}\makebox(0,0)[t]{\lineheight{1.25}\smash{\begin{tabular}[t]{c}\scalebox{0.6}{Mask}\end{tabular}}}}%
    \put(0.52653982,0.5158752){\color[rgb]{0.61568627,0,1}\makebox(0,0)[t]{\lineheight{1.25}\smash{\begin{tabular}[t]{c}\scalebox{0.6}{\textit{Pick up the pallet with concrete block on top}}\end{tabular}}}}%
  \end{picture}%
\endgroup%

%% file: figures/visualization_seg.pdf_tex
\begingroup%
  \makeatletter%
  \providecommand\color[2][]{%
    \errmessage{(Inkscape) Color is used for the text in Inkscape, but the package 'color.sty' is not loaded}%
    \renewcommand\color[2][]{}%
  }%
  \providecommand\transparent[1]{%
    \errmessage{(Inkscape) Transparency is used (non-zero) for the text in Inkscape, but the package 'transparent.sty' is not loaded}%
    \renewcommand\transparent[1]{}%
  }%
  \providecommand\rotatebox[2]{#2}%
  \newcommand*\fsize{\dimexpr\f@size pt\relax}%
  \newcommand*\lineheight[1]{\fontsize{\fsize}{#1\fsize}\selectfont}%
  \ifx\svgwidth\undefined%
    \setlength{\unitlength}{643.2021494bp}%
    \ifx\svgscale\undefined%
      \relax%
    \else%
      \setlength{\unitlength}{\unitlength * \real{\svgscale}}%
    \fi%
  \else%
    \setlength{\unitlength}{\svgwidth}%
  \fi%
  \global\let\svgwidth\undefined%
  \global\let\svgscale\undefined%
  \makeatother%
  \begin{picture}(1,0.6476064)%
    \lineheight{1}%
    \setlength\tabcolsep{0pt}%
    \put(0,0){\includegraphics[width=\unitlength,page=1]{visualization_seg.pdf}}%
    \put(0.02442587,0.33463292){\color[rgb]{0,0,0}\makebox(0,0)[t]{\lineheight{1.25}\smash{\begin{tabular}[t]{c}\scalebox{0.7}{\textit{a)}}\end{tabular}}}}%
    \put(0.24176239,0.33419245){\color[rgb]{0.61568627,0,1}\makebox(0,0)[t]{\lineheight{1.25}\smash{\begin{tabular}[t]{c}\scalebox{0.7}{\textit{Pick up the pallet}}\end{tabular}}}}%
    \put(0.56382334,0.33773727){\color[rgb]{0,0,0}\makebox(0,0)[t]{\lineheight{1.25}\smash{\begin{tabular}[t]{c}\scalebox{0.7}{\textit{b)}}\end{tabular}}}}%
    \put(0.79215737,0.33597839){\color[rgb]{0.61568627,0,1}\makebox(0,0)[t]{\lineheight{1.25}\smash{\begin{tabular}[t]{c}\scalebox{0.7}{\textit{Pick up the pallet covered by snow}}\end{tabular}}}}%
    \put(0.02953127,0.00739043){\color[rgb]{0,0,0}\makebox(0,0)[t]{\lineheight{1.25}\smash{\begin{tabular}[t]{c}\scalebox{0.7}{\textit{c)}}\end{tabular}}}}%
    \put(0.2626124,0.00703641){\color[rgb]{0.61568627,0,1}\makebox(0,0)[t]{\lineheight{1.25}\smash{\begin{tabular}[t]{c}\scalebox{0.7}{\textit{Pick up the pallet with a higher box}}\end{tabular}}}}%
    \put(0.56651377,0.00740493){\color[rgb]{0,0,0}\makebox(0,0)[t]{\lineheight{1.25}\smash{\begin{tabular}[t]{c}\scalebox{0.7}{\textit{d)}}\end{tabular}}}}%
    \put(0.79349695,0.00697952){\color[rgb]{0.61568627,0,1}\makebox(0,0)[t]{\lineheight{1.25}\smash{\begin{tabular}[t]{c}\scalebox{0.7}{\textit{Pick up the pallet with a package}}\end{tabular}}}}%
    \put(0,0){\includegraphics[width=\unitlength,page=2]{visualization_seg.pdf}}%
  \end{picture}%
\endgroup%

%% file: figures/pallet_accuracy_fp.pdf_tex
\begingroup%
  \makeatletter%
  \providecommand\color[2][]{%
    \errmessage{(Inkscape) Color is used for the text in Inkscape, but the package 'color.sty' is not loaded}%
    \renewcommand\color[2][]{}%
  }%
  \providecommand\transparent[1]{%
    \errmessage{(Inkscape) Transparency is used (non-zero) for the text in Inkscape, but the package 'transparent.sty' is not loaded}%
    \renewcommand\transparent[1]{}%
  }%
  \providecommand\rotatebox[2]{#2}%
  \newcommand*\fsize{\dimexpr\f@size pt\relax}%
  \newcommand*\lineheight[1]{\fontsize{\fsize}{#1\fsize}\selectfont}%
  \ifx\svgwidth\undefined%
    \setlength{\unitlength}{419.76194847bp}%
    \ifx\svgscale\undefined%
      \relax%
    \else%
      \setlength{\unitlength}{\unitlength * \real{\svgscale}}%
    \fi%
  \else%
    \setlength{\unitlength}{\svgwidth}%
  \fi%
  \global\let\svgwidth\undefined%
  \global\let\svgscale\undefined%
  \makeatother%
  \begin{picture}(1,0.86000228)%
    \lineheight{1}%
    \setlength\tabcolsep{0pt}%
    \put(0,0){\includegraphics[width=\unitlength,page=1]{pallet_accuracy_fp.pdf}}%
    \put(0.26711393,0.00926013){\makebox(0,0)[t]{\lineheight{1.25}\smash{\begin{tabular}[t]{c}\scalebox{0.9}{empty\_pallet}\end{tabular}}}}%
    \put(0.23165203,0.05435957){\makebox(0,0)[t]{\lineheight{1.25}\smash{\begin{tabular}[t]{c}\scalebox{0.9}{$e_x$[m]}\end{tabular}}}}%
    \put(0.13660244,0.1193746){\makebox(0,0)[t]{\lineheight{1.25}\smash{\begin{tabular}[t]{c}\scalebox{0.7}{0.00}\end{tabular}}}}%
    \put(0.06938107,0.18761958){\makebox(0,0)[t]{\lineheight{1.25}\smash{\begin{tabular}[t]{c}\scalebox{0.7}{4}\end{tabular}}}}%
    \put(0.02777161,0.27932546){\rotatebox{90.57785343}{\makebox(0,0)[t]{\lineheight{1.25}\smash{\begin{tabular}[t]{c}\scalebox{0.9}{$d_{\text{cam}}$[m]}\end{tabular}}}}}%
    \put(0.96179428,0.27506895){\rotatebox{-89.64164202}{\makebox(0,0)[t]{\lineheight{1.25}\smash{\begin{tabular}[t]{c}\scalebox{0.9}{$0^\circ-15^\circ$}\end{tabular}}}}}%
    \put(0.96187155,0.51185445){\rotatebox{-90.70705731}{\makebox(0,0)[t]{\lineheight{1.25}\smash{\begin{tabular}[t]{c}\scalebox{0.9}{$15^\circ-45^\circ$}\end{tabular}}}}}%
    \put(0.9606623,0.74720582){\rotatebox{-90.23330969}{\makebox(0,0)[t]{\lineheight{1.25}\smash{\begin{tabular}[t]{c}\scalebox{0.9}{$45^\circ-90^\circ$}\end{tabular}}}}}%
    \put(0.02869358,0.51442129){\rotatebox{90.57785343}{\makebox(0,0)[t]{\lineheight{1.25}\smash{\begin{tabular}[t]{c}\scalebox{0.9}{$d_{\text{cam}}$[m]}\end{tabular}}}}}%
    \put(0.02925099,0.74314224){\rotatebox{90.57785343}{\makebox(0,0)[t]{\lineheight{1.25}\smash{\begin{tabular}[t]{c}\scalebox{0.9}{$d_{\text{cam}}$[m]}\end{tabular}}}}}%
    \put(0.06888224,0.27439207){\makebox(0,0)[t]{\lineheight{1.25}\smash{\begin{tabular}[t]{c}\scalebox{0.7}{6}\end{tabular}}}}%
    \put(0.06954823,0.36089512){\makebox(0,0)[t]{\lineheight{1.25}\smash{\begin{tabular}[t]{c}\scalebox{0.7}{8}\end{tabular}}}}%
    \put(0.0695979,0.4187631){\makebox(0,0)[t]{\lineheight{1.25}\smash{\begin{tabular}[t]{c}\scalebox{0.7}{4}\end{tabular}}}}%
    \put(0.06931598,0.50627211){\makebox(0,0)[t]{\lineheight{1.25}\smash{\begin{tabular}[t]{c}\scalebox{0.7}{6}\end{tabular}}}}%
    \put(0.0692284,0.59379022){\makebox(0,0)[t]{\lineheight{1.25}\smash{\begin{tabular}[t]{c}\scalebox{0.7}{8}\end{tabular}}}}%
    \put(0.06942152,0.65071743){\makebox(0,0)[t]{\lineheight{1.25}\smash{\begin{tabular}[t]{c}\scalebox{0.7}{4}\end{tabular}}}}%
    \put(0.06869614,0.73822807){\makebox(0,0)[t]{\lineheight{1.25}\smash{\begin{tabular}[t]{c}\scalebox{0.7}{6}\end{tabular}}}}%
    \put(0.06922407,0.82540527){\makebox(0,0)[t]{\lineheight{1.25}\smash{\begin{tabular}[t]{c}\scalebox{0.7}{8}\end{tabular}}}}%
    \put(0.24408047,0.12005434){\makebox(0,0)[t]{\lineheight{1.25}\smash{\begin{tabular}[t]{c}\scalebox{0.7}{0.25}\end{tabular}}}}%
    \put(0.35622705,0.11970846){\makebox(0,0)[t]{\lineheight{1.25}\smash{\begin{tabular}[t]{c}\scalebox{0.7}{0.5}\end{tabular}}}}%
    \put(0.44560321,0.11744144){\makebox(0,0)[t]{\lineheight{1.25}\smash{\begin{tabular}[t]{c}\scalebox{0.7}{0.25}\end{tabular}}}}%
    \put(0.51418526,0.11809345){\makebox(0,0)[t]{\lineheight{1.25}\smash{\begin{tabular}[t]{c}\scalebox{0.7}{0.00}\end{tabular}}}}%
    \put(0.59379746,0.11885656){\makebox(0,0)[t]{\lineheight{1.25}\smash{\begin{tabular}[t]{c}\scalebox{0.7}{0.25}\end{tabular}}}}%
    \put(0.75340605,0.12135626){\makebox(0,0)[t]{\lineheight{1.25}\smash{\begin{tabular}[t]{c}\scalebox{0.7}{0.0}\end{tabular}}}}%
    \put(0.84523528,0.11840199){\makebox(0,0)[t]{\lineheight{1.25}\smash{\begin{tabular}[t]{c}\scalebox{0.7}{0.1}\end{tabular}}}}%
    \put(0.93599257,0.11990738){\makebox(0,0)[t]{\lineheight{1.25}\smash{\begin{tabular}[t]{c}\scalebox{0.7}{0.2}\end{tabular}}}}%
    \put(0.80596766,0.05366171){\makebox(0,0)[t]{\lineheight{1.25}\smash{\begin{tabular}[t]{c}\scalebox{0.9}{$e_z$[m]}\end{tabular}}}}%
    \put(0.52211893,0.05416369){\makebox(0,0)[t]{\lineheight{1.25}\smash{\begin{tabular}[t]{c}\scalebox{0.9}{$e_y^{\psi}$[m]}\end{tabular}}}}%
    \put(0.54947425,0.00901195){\makebox(0,0)[t]{\lineheight{1.25}\smash{\begin{tabular}[t]{c}\scalebox{0.9}{big\_box}\end{tabular}}}}%
    \put(0.7920569,0.00869354){\makebox(0,0)[t]{\lineheight{1.25}\smash{\begin{tabular}[t]{c}\scalebox{0.9}{small\_box}\end{tabular}}}}%
    \put(0,0){\includegraphics[width=\unitlength,page=2]{pallet_accuracy_fp.pdf}}%
  \end{picture}%
\endgroup%

%% file: 5_conclusions.tex
\section{Conclusions}\label{Sec:con}
  
Lang2Lift demonstrates the practical integration of language-guided perception and pose estimation for autonomous forklift operation in outdoor industrial environments.
The experimental evaluation further highlights practical lessons for deploying language-guided perception in industrial automation systems. Observed failure modes primarily arise from ambiguous natural language instructions and severe occlusions, underscoring the importance of clear operator command protocols and conservative perception validation during execution. From a deployment perspective, the results indicate that foundation-model-based perception can be integrated into forklift operations when paired with tolerance-aware pose processing, temporal tracking, and low-speed control strategies. These findings provide a concrete engineering roadmap for practitioners seeking to incorporate language-guided perception into outdoor material-handling systems, while outlining clear directions for incremental improvements, such as sensor redundancy, perception latency reduction, and system-level robustness, rather than algorithmic novelty.